\documentclass[10pt,twocolumn,letterpaper]{article}

\usepackage{cvpr}              
\newcommand{\ignore}[1]{}

\newcommand{\ba}{\begin{array}}
        \newcommand{\ea}{\end{array}}
\newcommand{\bc}{\begin{center}}
        \newcommand{\ec}{\end{center}}
\newcommand{\be}{\begin{enumerate}}
        \newcommand{\ee}{\end{enumerate}}
\newcommand{\bea}{\begin{eqnarray}}
        \newcommand{\eea}{\end{eqnarray}}
\newcommand{\beas}{\begin{eqnarray*}}
        \newcommand{\eeas}{\end{eqnarray*}}
\newcommand{\beq}{\begin{equation}}
        \newcommand{\eeq}{\end{equation}}
\newcommand{\bfig}{\begin{figure}}
        \newcommand{\efig}{\end{figure}}
\newcommand{\bi}{\begin{itemize}}
        \newcommand{\ei}{\end{itemize}}
\newcommand{\bpic}{\begin{picture}}
        \newcommand{\epic}{\end{picture}}
\newcommand{\btabular}{\begin{tabular}}
        \newcommand{\etabular}{\end{tabular}}
\newcommand{\btable}{\begin{table}}
        \newcommand{\etable}{\end{table}}

\newcommand{\es}{\vfill
    \rule[-6mm]{170mm}{0.7mm} \\
    \redw{{\tiny
                \hfill S-\theslide}}
    \end{slide}}

\newcommand{\matxx}[1]{{\mathbf #1}}
\newcommand{\vecXX}[1]{{\mathbf {#1}}}
\newcommand{\vecYY}[1]{{\boldsymbol {#1}}}

\newcommand{\argmin}{\operatornamewithlimits{arg\ min}}

\def \hbar {{\bar{h}}}

\def \vecm {{\vecXX{m}}}

\def \vecq {{\vecXX{q}}}

\def \vect {{\vecXX{t}}}

\def \vecx {{\vecXX{x}}}

\def \matC {{\matxx{C}}}
\def \matD {{\matxx{D}}}

\def \matQ {{\matxx{Q}}}
\def \matR {{\matxx{R}}}

\def \matT {{\matxx{T}}}

\def \matX {{\matxx{X}}}

\newcommand{\mattwo}[4]{\left[\begin{array}{cc}#1&#2\\#3&#4\end{array}\right]}

\newcommand{\pdiff}[2]{ \frac{\partial#1}{\partial#2} }

\makeatletter
\renewcommand*\env@matrix[1][*\c@MaxMatrixCols c]{%
    \hskip -\arraycolsep
    \let\@ifnextchar\new@ifnextchar
    \array{#1}}
\makeatother

\newcommand{\RR}{\mathbb{R}}

\newcommand{\SO}[1]{\ensuremath{\mathbf{SO}(#1)}}

\newcommand{\Sim}[1]{\ensuremath{\mathbf{Sim}(#1)}}

\newcommand{\simlie}[1]{\ensuremath{{\mathfrak{sim}(#1)}}}

\font\Bigmath=cmsy10 scaled \magstep2
\def\dplus{\mathrel{%
        \ooalign{$+$\cr\hss\lower.255ex\hbox{\Bigmath\char5}\hss}}}
\def\dminus{\mathrel{%
        \ooalign{$-$\cr\hss\lower.255ex\hbox{\Bigmath\char5}\hss}}}

\newcommand{\cD}{\mathcal{D}}

\newcommand{\cF}{\mathcal{F}}

\newcommand{\cM}{\mathcal{M}}

\DeclareMathOperator*{\Exp}{Exp}

\def\bea#1\eea{\begin{align}#1\end{align}}
\def\beas#1\eeas{\begin{align*}#1\end{align*}}

\usepackage{microtype}

\usepackage{xspace}

\usepackage{multirow} 
\usepackage{boldline}

\newcommand{\duster}{DUSt3R\xspace}
\newcommand{\master}{MASt3R\xspace}
\newcommand{\mastersfm}{MASt3R-SfM\xspace}
\newcommand{\spanner}{Spann3R\xspace}

\newcommand{\Fmaster}{\cF_{M}}

\newcommand{\mbf}[1]{\mathbf{#1}}
\newcommand{\mcal}[1]{\mathcal{#1}}

\newcommand{\normsq}[1]{\left\Vert#1\right\Vert^2}
\newcommand{\ray}[1]{\psi\left(#1\right)}
\newcommand{\projection}{\Pi}
\newcommand{\proj}[1]{\projection\left(#1\right)}

\newcommand{\match}[2]{\vecm_{#1,#2}}
\newcommand{\matchconf}[2]{\vecq_{#1,#2}}

\newcommand{\J}{\mbf{J}}

\newcommand{\I}{\mcal{I}}
\newcommand{\Ii}{\I^i}
\newcommand{\Ij}{\I^j}
\newcommand{\Ik}{\I^k}
\newcommand{\If}{\I^f}

\newcommand{\pix}{\mbf{p}}

\newcommand{\X}{\matX}
\newcommand{\Xii}{\X^i_{i}}
\newcommand{\Xji}{\X^j_{i}}

\newcommand{\Xij}{\X^i_{j}}
\newcommand{\Xkk}{\X^k_k}
\newcommand{\Xkf}{\X^k_f}

\newcommand{\Xff}{\X^f_f}

\newcommand{\Xffm}{\X^f_{f, m}}

\newcommand{\Xcanon}{\tilde{\matX}}
\newcommand{\Xcanonii}{\Xcanon^i_{i}}
\newcommand{\Xcanonkk}{\Xcanon^k_{k}}
\newcommand{\Xcanonkkn}{\Xcanon^k_{k,n}}
\newcommand{\Xcanoniim}{\Xcanon^i_{i,m}}
\newcommand{\Xcanonjjn}{\Xcanon^j_{j,n}}

\newcommand{\Ccanon}{\tilde{\matC}}
\newcommand{\Ccanonkk}{\Ccanon^k_{k}}

\newcommand{\C}{\matC}
\newcommand{\Cii}{\C^i_{i}}
\newcommand{\Cji}{\C^j_{i}}

\newcommand{\Ckf}{\C^k_f}

\newcommand{\Q}{\matQ}
\newcommand{\Qii}{\Q^i_{i}}
\newcommand{\Qji}{\Q^j_{i}}

\newcommand{\D}{\matD}
\newcommand{\Dii}{\D^i_{i}}
\newcommand{\Dji}{\D^j_{i}}

\newcommand{\Twci}{\matT_{WC_i}}
\newcommand{\Twcj}{\matT_{WC_j}}
\newcommand{\Tkf}{\matT_{kf}}

\newcommand{\Tij}{\matT_{ij}}

\newcommand{\vectau}{\vecYY{\tau}}
\newcommand{\KF}{\mathcal{K}}
\newcommand{\KFthreshold}{\omega_k}
\newcommand{\LCthreshold}{\omega_l}
\newcommand{\Retrievalthreshold}{\omega_r}

\newcommand{\kdtree}{$k$-d tree\xspace}

\newcommand\blfootnote[1]{%
  \begingroup
  \renewcommand\thefootnote{}\footnote{#1}%
  \addtocounter{footnote}{-1}%
  \endgroup
}

\definecolor{cvprblue}{rgb}{0.21,0.49,0.74}
\usepackage[pagebackref,breaklinks,colorlinks,allcolors=cvprblue]{hyperref}

\def\paperID{10509} 
\def\confName{CVPR}
\def\confYear{2025}

\title{MASt3R-SLAM: Real-Time Dense SLAM with 3D Reconstruction Priors}

\author{
    \makebox[0.32\textwidth][r]{Riku Murai\textsuperscript{*}} 
    \makebox[0.32\textwidth][c]{Eric Dexheimer\textsuperscript{*}} 
    \makebox[0.32\textwidth][l]{Andrew J. Davison} \\ 
    Imperial College London\\
    {\tt\small \{riku.murai15, e.dexheimer21, a.davison\}@imperial.ac.uk} \\
}

\begin{document}
\maketitle
\begin{abstract}
We present a real-time monocular dense SLAM system designed bottom-up from MASt3R, a two-view 3D reconstruction and matching prior.  Equipped with this strong prior, our system is robust on in-the-wild video sequences despite making no assumption on a fixed or parametric camera model beyond a unique camera centre.  We introduce efficient methods for pointmap matching, camera tracking and local fusion, graph construction and loop closure, and second-order global optimisation. With known calibration, a simple modification to the system achieves state-of-the-art performance across various benchmarks.  Altogether, we propose a plug-and-play monocular SLAM system capable of producing globally consistent poses and dense geometry while operating at 15 FPS.
\end{abstract}
{\blfootnote{*Authors contributed equally to this work.}}
\section{Introduction}

Visual simultaneous localisation and mapping (SLAM) is a foundational building block for today's robotics and augmented reality products.  
With careful design of an integrated hardware and software stack, robust and accurate visual SLAM is now possible.  
However, SLAM is not yet a plug-and-play algorithm as it requires hardware expertise and calibration.  
For a minimal single camera setup without additional sensing such as an IMU, in-the-wild SLAM that provides both accurate poses and consistent dense maps does not exist.  Achieving such a reliable dense SLAM system would open new research avenues for spatial intelligence.  

Performing dense SLAM from only 2D images requires reasoning over time-varying  poses and camera models, as well as 3D scene geometry.  To solve such an inverse problem of large dimensionality, a variety of priors, from handcrafted to data-driven, have been proposed.  Single-view priors, such as monocular depth and normals, attempt to predict geometry from a single image, but these contain ambiguities and lack consistency across views.  
While multi-view priors like optical flow reduce ambiguity, decoupling pose and geometry is challenging since pixel motion depends on both the extrinsics and the camera model. Although these underlying causes may vary across time and different observers, the 3D scene remains invariant across views.
Therefore, the unifying prior required to solve for poses, camera models, and dense geometry from images is over the space of 3D geometry in a common coordinate frame. 
\label{sec:intro}
\begin{figure}[t]
	\centering
	\includegraphics[width=0.98\linewidth]{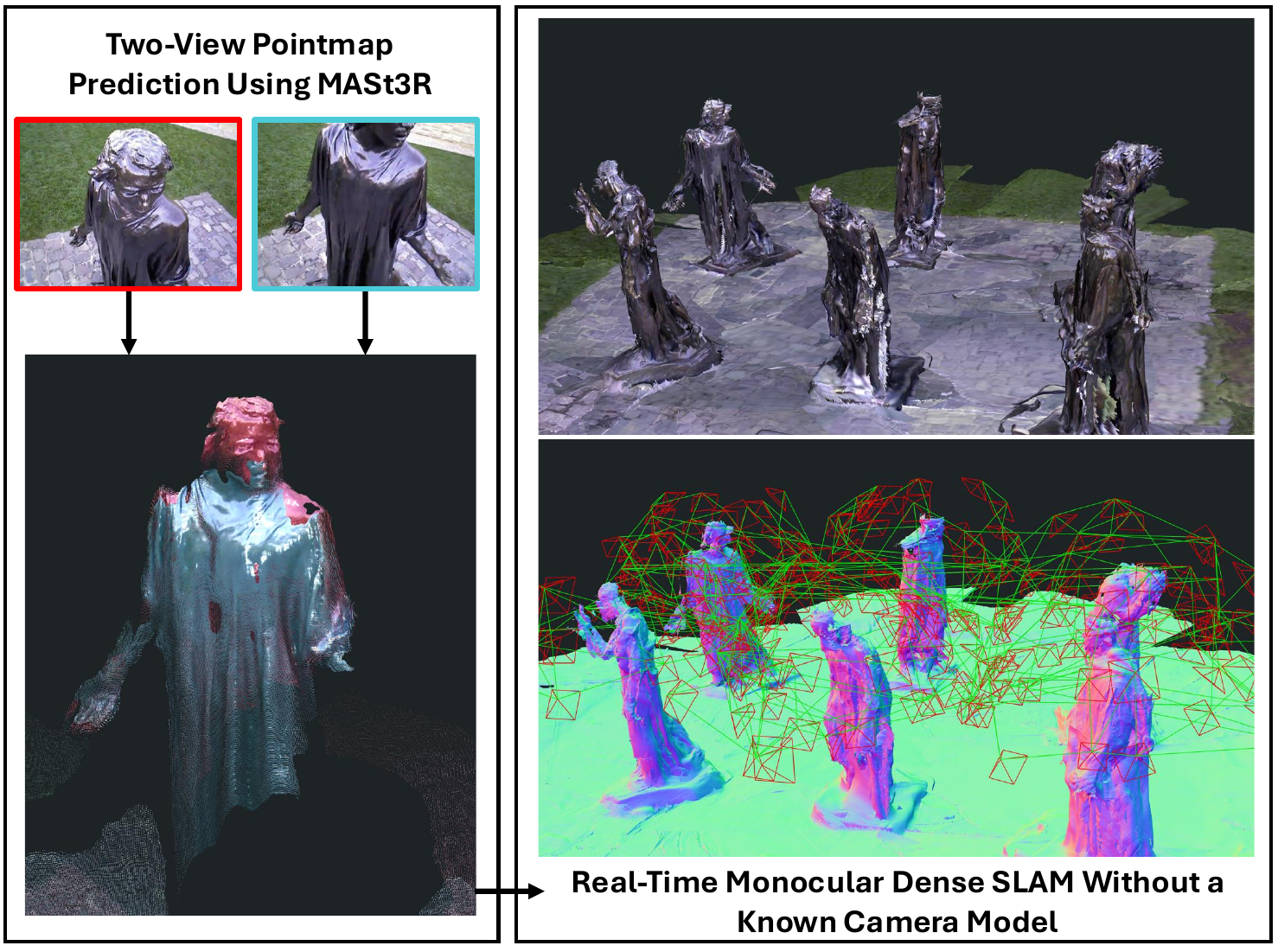}
	\caption{Reconstruction from our dense monocular SLAM system on the Burghers sequence \cite{zhou_burghers_2013}.  Using two-view predictions from \master shown on the left, our system achieves globally consistent poses and geometry in real-time without a known camera model.}
	\label{fig:teaser} 
\end{figure}

Recently, \textit{two-view 3D reconstruction priors}, pioneered by \duster \cite{wang_dust3r_cvpr24} and its successor \master \cite{leroy_mast3r_eccv24}, have created a paradigm shift in structure-from-motion (SfM) by capitalising on curated 3D datasets.  These networks output pointmaps directly from two images in a common coordinate frame, such that the aforementioned subproblems are implicitly solved in a joint framework.  In the future, these priors will be trained on all varieties of camera models with significant distortion.  While 3D priors could take in more views, SfM and SLAM leverage spatial sparsity and avoid redundancy to achieve large-scale consistency.  A two-view architecture mirrors two-view geometry as the building block of SfM, and this modularity opens the door for both efficient decision-making and robust consensus in the backend.  

In this work, we propose the first real-time SLAM framework to leverage two-view 3D reconstruction priors as a unifying foundation for tracking, mapping, and relocalisation as shown in \cref{fig:teaser}.  While previous work has applied these priors to SfM in an offline setting with unordered image collections \cite{duisterhof_2024_mast3rsfm}, SLAM receives data incrementally and must maintain real-time operation.  This requires new perspectives on low-latency matching, careful map maintenance, and efficient methods for large-scale optimisation.  Furthermore, inspired by both filtering and optimisation techniques in SLAM, we perform local filtering of pointmaps in the frontend to enable large-scale global optimisation in the backend.  Our system makes no assumption on each image's camera model beyond having a unique camera centre that all rays pass through.  This results in a real-time dense monocular SLAM system capable of reconstructing scenes with generic, time-varying camera models.  Given calibration, we also demonstrate state-of-the-art performance in trajectory accuracy and dense geometry estimation.

In summary, our contributions are:
\begin{itemize}
    \item The first real-time SLAM system using the two-view 3D reconstruction prior \master \cite{leroy_mast3r_eccv24} as a foundation.
    \item Efficient techniques for pointmap matching, tracking and local fusion, graph construction and loop closure, and second-order global  optimisation.
    \item A state-of-the-art dense SLAM system capable of handling generic, time-varying camera models.
\end{itemize}

\section{Related Work}
\label{sec:related_work}

To obtain accurate pose estimation, \textbf{sparse monocular SLAM} focuses on jointly solving for camera poses and a select number of unbiased 3D landmarks \cite{davison_monoslam_2007}.  Algorithmic advances leveraging the sparsity of the optimisation \cite{klein_parallel_2007} and careful graph construction \cite{mur-artal_orb-slam_2015} enabled real-time pose estimation and sparse reconstructions on large scale scenes.  While sparse monocular SLAM is very accurate given sufficient features and parallax, it lacks a dense scene model which is useful for both robust tracking and more explicit reasoning over geometry.

To improve robustness and provide interaction, early \textbf{dense monocular SLAM} systems demonstrated alternating optimisation of poses and dense depth with handcrafted regularisation \cite{newcombe_dtam_2011}.  As these systems were limited to controlled settings, recent work has attempted to combine data-driven priors with backend optimisation.  While predicting geometric quantities from a single image, such as depth \cite{eigen_depth_2014, ranftl_midas_2022, ke_marigold_2024, yang_depthanything_2024} and surface normals \cite{Wang_2015_CVPR, bae2024dsine}, have shown significant progress, their use has been limited in SLAM. Predicting geometry from a single-view is ambiguous, resulting in biased and inconsistent 3D geometry.  SLAM literature has thus focused on predicting priors over a hypothesis space of possible depths in the form of latent spaces \cite{bloesch_codeslam_2018, czarnowski_deepfactors_2020}, subspaces \cite{tang_ba-net_2019}, local primitives \cite{mazur_superprimitive_2024}, and distributions \cite{dexheimer_depthcov_2023, dexheimer_como_2024}.  While the flexibility of these priors can achieve greater consistency, robust correspondence across multiple views is essential.

\textbf{Multi-view priors}, such as multi-view stereo (MVS) \cite{zhou_deeptam_2018, koestler_tandem_2022, sayed2022simplerecon} and optical flow \cite{teed_raft_2020}, instead focus on learning correspondence from two or more views as a means to obtaining geometry.  However, both require additional information: MVS fixes poses to achieve correspondence, while flow is an entangled observation of motion and geometry subject to the degeneracies mentioned previously.  DROID-SLAM \cite{teed_droid_2021} combines learned features for matching along with a per-pixel dense bundle adjustment framework into a single end-to-end framework.  
This results in a robust SLAM system with a backend similar in spirit to sparse SLAM, so the lack of explicit geometric constraints can still produce inconsistent 3D geometry.

\textbf{Volumetric representations} have demonstrated the potential for consistent reconstruction as geometry parameters are coupled in the rendering process.  A variety of SLAM systems have adopted differentiable rendering in neural fields \cite{mildenhall_nerf_2020} and Gaussian splatting \cite{kerbl_3d_2023} for both monocular \cite{zhu_nicer_2024, matsuki_gaussian_2024} and RGB-D \cite{sucar_imap_2021, zhu_nice_2022, keetha_splatam_2024, yan2024gs} cameras.  However, these methods have lagged in real-time performance compared to alternatives, and require depth, additional 2D priors, or slow camera motion to constrain the solution.  3D priors for general scene reconstruction from images first fuse 2D features into 3D voxel grids which are then decoded into surface geometry \cite{murez2020atlas, sun2021neucon}.  These methods assume known poses for fusion, so are unsuitable for joint tracking and mapping, while the volumetric representations require significant memory and a pre-defined resolution.

All systems mentioned thus far assume known intrinsic calibration. Classical \textbf{automatic intrinsic calibration} is possible when there are strict assumptions on scene geometry or unchanging parameters across a set of images \cite{hartley2003multiple}, but encounters degenerate configurations and sensitivity to noise.  Given an initial estimate of intrinsics, refinement via bundle adjustment can improve accuracy online \cite{nima_constant_2014}, but this already assumes a parametric model and sufficient initialisation of all parameters.  Combining DROID-SLAM and self-calibration \cite{hagemann2023deep} yields improved robustness to noisy intrinsics, but optimisation is slower due to denser matrix fill-in.  More recently, data-driven methods predict intrinsics from one or multiple images \cite{jin2023perspective, veicht2024geocalib}, but are either limited in accuracy for in-the-wild SLAM or are not flexible in the camera model definition.   

Recently, \duster introduced a novel two-view \textbf{3D \nobreak reconstruction prior} that outputs dense 3D point clouds of both images in a \textit{common} coordinate frame.  Compared to previously discussed priors that solve subproblems of the task, \duster provides a direct pseudo-measurement of a two-view 3D scene by implicitly reasoning over correspondence, poses, camera models, and dense geometry.  The successor \master \cite{leroy_mast3r_eccv24} predicts additional per-pixel features to improve pixel matching for localisation and SfM \cite{duisterhof_2024_mast3rsfm}.  However, as with all priors, predictions can still have inconsistencies and correlated errors in the 3D geometry. \duster and \mastersfm thus require large-scale optimisation for global consistency, but the time complexity does not scale well with the number of images.  \spanner \cite{wang2024spann3r} forgoes backend optimisation by fine-tuning \duster to predict a stream of pointmaps directly into a global coordinate system, but must maintain a limited memory of tokens which can cause drift in larger scenes.

In this work, we propose a dense SLAM system built around these two-view 3D reconstruction priors.  We only assume a generic central camera model, and propose efficient methods for pointmap matching, tracking and pointmap fusion, loop closure, and global optimisation to achieve large scale consistency of the pairwise predictions in real-time.

\section{Method}
\label{sec:method}

\begin{figure*}[ht]
\centering
\includegraphics[width=\linewidth]{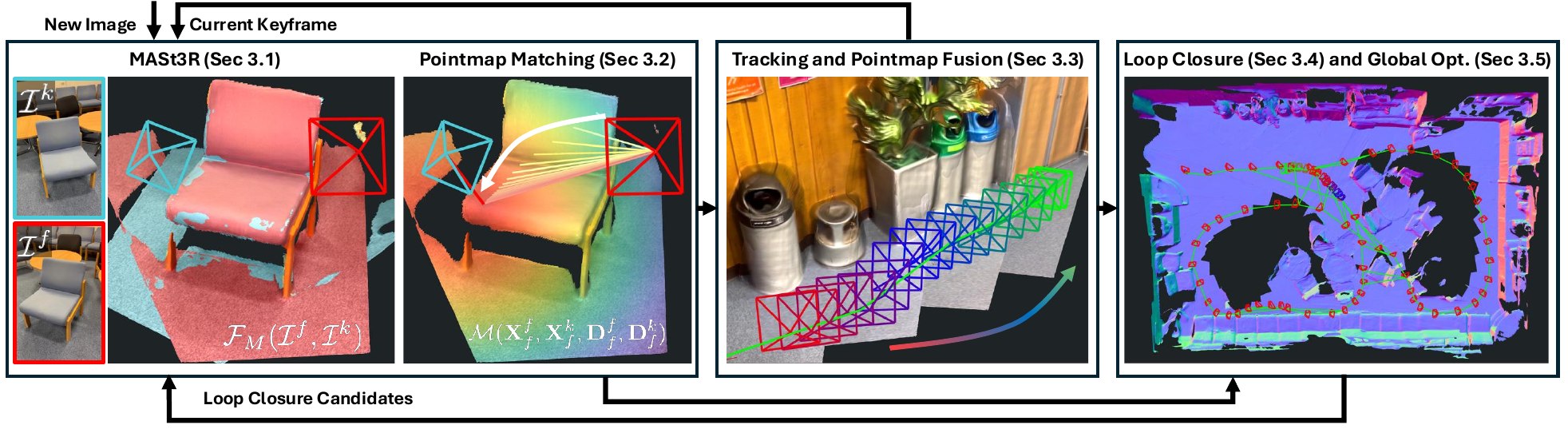}
\caption{System diagram of \master-SLAM.  New images are tracked against the current keyframe by predicting a pointmap from \master and finding pixel matches using our efficient iterative projection pointmap matching.  Tracking estimates the current pose and performs local pointmap fusion.  When new keyframes are added to the backend, loop closure candidates are selected by querying the retrieval database using encoded \master features.  Candidates are then decoded by \master and if a sufficient number of matches is found, edges are added to the backend graph. Large-scale second-order optimisation achieves global consistency of poses and dense geometry.}
\label{fig:system_diagram}
\end{figure*}

We provide an overview of the method in \cref{fig:system_diagram}, which shows our main components: \master prediction and pointmap matching, tracking and local fusion, loop closure, and global optimisation.

\subsection{Preliminaries}
\duster takes in a pair of images $\Ii, \Ij \in \RR^{H\times W\times3}$, and outputs pointmaps $\Xii, \Xji \in \RR^{H\times W\times3}$ along with their confidences $\Cii, \Cji \in \RR^{H\times W\times1}$. Here, we use notation $\Xij$ to express the pointmap of image $i$ represented in the coordinate frame of camera $j$.
In \master, an additional head is added to predict $d$-dimensional features for matching $\Dii, \Dji\in \RR^{H\times W\times d}$ and its corresponding confidences $\Qii, \Qji\in \RR^{H\times W\times1}$.  We define $\Fmaster(\Ii, \Ij)$ as the forward pass of \master that yields the previously discussed outputs, and throughout the text we will use \master's output directly for conciseness.

While some of the data used to train \master has metric scale, we found that scale is often a large source of inconsistency across predictions. 
To optimise over differently scaled predictions, we define all poses as $\matT \in \Sim{3}$ and updates to the poses using Lie algebra $\vectau \in \simlie{3}$ and a left-plus operator: 
\beq
\matT = \mattwo{s\matR}{\vect}{0}{1}~, \quad
\matT \leftarrow \vectau \oplus \matT \triangleq \Exp(\vectau) \circ \matT,
\eeq
where $\matR\in\SO{3}$, $\vect \in \RR^3$, and scale $s \in \RR$, following the notation in \cite{sola_micro_2018, teed2021tangent}.

Our only assumption on the camera model is that of a generic central camera \cite{schops_why_2020}, which means that all rays pass through a unique camera centre.  We define the function $\ray{\Xii}$ that normalises a pointmap $\Xii$ into rays of unit norm such that each pointmap defines its own camera model.  This enables handling both time-varying camera models, such as zoom, and distortion in a unified manner.

\subsection{Pointmap Matching}\label{sec:matching}
Correspondence is a fundamental component of SLAM that is required for both tracking and mapping.  In this case, given the pointmaps and features from \master, we need to find the set of pixel matches between the two images, denoted by $\match{i}{j} = \cM(\X^i_{\smash{i}}, \X^j_{\smash{i}}, \D^i_{\smash{i}}, \D^j_{\smash{i}})$.  Naive brute-force matching has quadratic complexity since it is a global search over all possible pairs of pixels.  To avoid this, \duster uses a \kdtree over 3D points; however, construction is non-trivial to parallelise and the nearest-neighbour search in 3D will find many inaccurate matches if there are errors in the pointmap predictions.  
In \master, additional high-dimensional features are predicted from the network to achieve wider baseline matching and a coarse-to-fine scheme is proposed to handle the global search.  However, the runtime is on the order of seconds for dense pixel matching, and sparse matching is still slower than the \kdtree.  
Rather than focusing on efficient methods for a global search over matches, we instead find inspiration from optimisation as a local search. 

\begin{figure}[t]
	\centering
	\includegraphics[width=0.98\columnwidth]{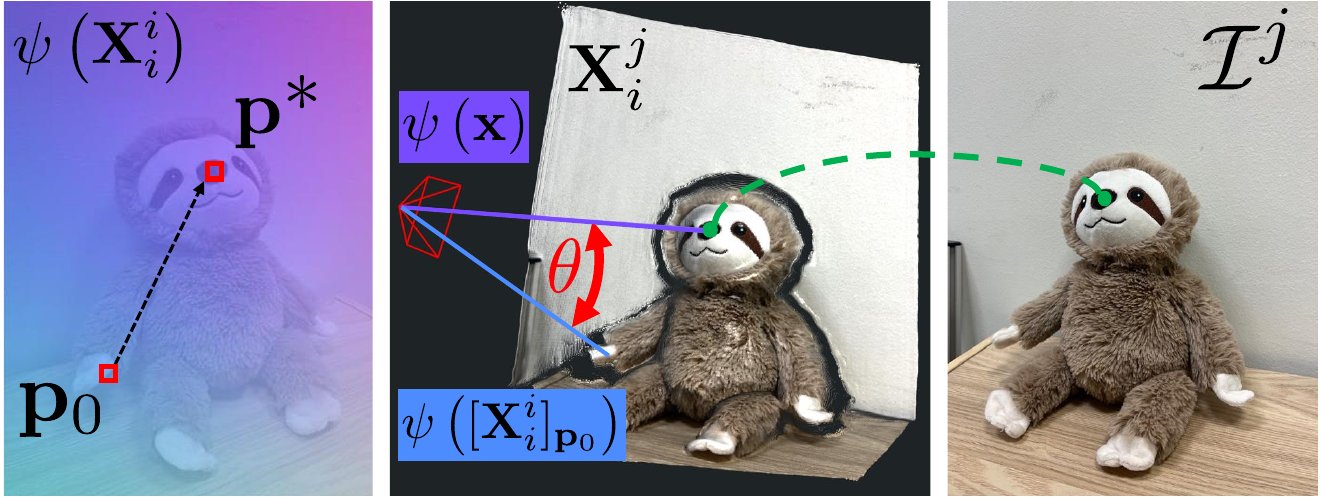}
	\caption{Overview of iterative projective matching: given the two pointmap predictions from \master, the reference pointmap is normalised $\ray{\Xii}$ to give a smooth pixel to ray mapping.  For an initial estimate of the projection $\pix_0$ of 3D point $\mbf{x}$ from pointmap $\Xji$, the pixel is iteratively updated to minimise the angular difference $\theta$ between the queried ray $\smash{\ray{[\Xii]_\pix}}$ and the target ray $\ray{\mbf{x}}$.  After finding the pixel $\pix^*$ that achieves the minimum error, we have a pixel correspondence between $\Ii$ and $\Ij$.}
	\label{fig:matching} 
\end{figure}

Compared to feature matching, we are motivated by the use of projective data-association commonly used in dense SLAM.  However, this requires a parametric camera model with closed-form projection, while our only assumption is that each frame has a unique camera centre.  Given the output pointmaps $\Xii, \Xji$, we can construct the generic camera model of $\Ii$ with the rays 
$\ray{\Xii}$.  Inspired by generic camera calibration methods \cite{rosebrock_generic_2012, schops_why_2020} which lack closed-form projection, we project each point $\mbf{x} \in \Xji$ independently by iteratively optimising the pixel coordinates $\pix$ in the reference frame that minimise the ray error:
\beq
    \pix^* = \argmin_{\pix} \normsq{\ray{[\Xii]_\pix} - \ray{\mbf{x}}}.
\eeq
We show a visual overview in \cref{fig:matching}, and note that minimising the Euclidean distance between normalised vectors is equivalent to minimising the angle $\theta$ between two normalised rays:
\beq
    \normsq{\psi_1 - \psi_2} = 2(1-\cos{\theta}), \quad \cos{\theta} = \psi_1^T \psi_2.
    \label{eq:ray_error}
\eeq
By using the nonlinear least-squares form similar to \cite{schops_why_2020}, we can iteratively solve for updates to projected locations by calculating analytical Jacobians and solving via Levenberg-Marquardt.
This can be done separately for each point and converges for almost all valid pixels within 10 iterations as the ray image is smooth.  At the end of this process, we now have initial matches $\match{i}{j}$.  When there is no initial estimate for the projection $\pix$, such as when tracking against a new keyframe or when matching loop closure edges, all pixels are initialised with the identity mapping.  During tracking, since we always have the matches from the previous frame, we can use this as initialisation to further speed up the convergence.  To handle occlusions and outliers, we also invalidate matches that have large distances in 3D space.  Our matching is massively parallel on GPU and additionally can leverage the incremental nature of SLAM.

While these pixels give a good initial estimate of matches using the geometry, \master demonstrates that leveraging per-pixel features greatly improves downstream performance on pose estimation.  Since we have a good initialisation from the previous step, we conduct a coarse-to-fine image-based search by updating the pixel location to the maximum feature similarity in a local patch window.

We implement both the iterative projection and feature refinement steps in custom CUDA kernels, as both are parallelisable for each pixel. For tracking this takes only 2 milliseconds and for constructing edges in the graph this takes only a few milliseconds for all newly added edges without any initial estimates of the projections.  Note that our matches are unbiased by our pose estimates as they rely purely on the \master outputs, which is atypical for projective data association.

\subsection{Tracking and Pointmap Fusion}\label{sec:tracking}

A key component of SLAM is low-latency tracking of the current frame's pose against the map.  As a keyframe-based system, we estimate the relative transformation $\Tkf$ between the current frame $\If$ and the last keyframe $\Ik$.  To be efficient, we would like to use only a single pass of the network to estimate the transformation. Assuming we already have the last keyframe's pointmap estimate $\Xcanonkk$, we need points in the frame of $\If$ to resolve $\Tkf$. This can be obtained via $\Fmaster(\If, \Ik)$.
One straightforward method to solve for pose is minimising the 3D point error:
\beq
E_p = \sum_{{m, n} \in \match{f}{k}}\left\|\frac{\Xcanonkkn - \Tkf \Xffm}{w(\matchconf{m}{n}, \sigma^2_p)}\right\|_{\rho}~, \label{eq:track_point}
\eeq
where 
$\matchconf{m}{n}=\sqrt{\Q^f_{\smash{f,m}} \Q^k_{\smash{f,n}}}$ is the match confidence weighting proposed in \mastersfm~\cite{duisterhof_2024_mast3rsfm}. For robustness, in addition to the Huber norm $\|\cdot\|_\rho$, a per-match weighting is applied:
\beq
w(\vecq, \sigma^2) = 
\begin{cases}
{\sigma^2/\vecq} & \vecq > \vecq_{min} \\ 
\infty &\text{otherwise} \\
\end{cases} ~.
\eeq
While $\Xkf$ instead of $\Xff$ could also be aligned to $\Xkk$ with the benefit of no explicit matching required as they are pixel aligned, we found that explicit matching with $\Xff$ had improved accuracy for larger baseline scenarios.  More importantly, although the 3D point error is suitable, it is easily skewed by errors in the pointmap predictions as inconsistent predictions in depth are relatively frequent. Since we ultimately fuse predictions into a single pointmap that averages out all the predictions, error in tracking degrades the keyframe's pointmap that will also be used in the backend.  

By again exploiting that the pointmap predictions can be converted to rays under a central camera assumption,
we can calculate a directional ray error instead, which is less sensitive to incorrect depth predictions. To calculate this, we simply normalise both points from \cref{eq:track_point}:
\beq
E_r = \sum_{{m, n} \in \match{f}{k}}\left\|\frac{\ray{\Xcanonkkn} - \ray{\Tkf \Xffm}}{w(\matchconf{m}{n}, \sigma^2_r)}\right\|_{\rho}~. \label{eq:track_ray}
\eeq
This results in a similar angular error as mentioned in \cref{eq:ray_error} and shown in \cref{fig:matching}, except that we now have many known correspondences and wish to find the pose that minimises all angular errors between canonical rays and corresponding predicted rays from the current frame.  Since angular errors are bounded, ray-based errors are robust against outliers \cite{pan2024glomap}.  We also include an error term with a small weight on the difference in distances from the camera centre.
This prevents the system from becoming degenerate under pure rotation, while avoiding significant bias from errors in depth.  We efficiently solve for updates to the pose using Gauss-Newton in an iteratively reweighted least-squares (IRLS) framework.  We calculate analytical Jacobians of the ray and distance errors with respect to a perturbation $\vectau$ of the relative pose $\Tkf$.   
We stack the residuals, Jacobians, and weights into matrices $\mbf{r}$, $\J$, and $\mbf{W}$, respectively. We iteratively solve the linear system and update the pose via:
\beq
    \left(\J^T \mbf{W} \J\right) \vectau =  -\J^T \mbf{W} \mbf{r}, \quad
\Tkf \leftarrow \vectau \oplus \Tkf.
    \label{eq:gauss_newton}
\eeq
Since each pointmap may provide valuable new information, we leverage this by not only filtering over estimates of the geometry, but also over the camera model itself, since it is defined by the rays.  After solving for the relative pose, we can use transform $\Tkf$ and update the canonical pointmap $\Xcanonkk$ via a running weighted average filter \cite{curless_tsdf_1996, newcombe_kinectfusion_2011}:
\beq
\Xcanonkk \leftarrow \frac{\Ccanonkk \Xcanonkk +  \Ckf \left(\Tkf \Xkf\right)}{ \Ccanonkk + \Ckf}, 
\Ccanonkk \leftarrow \Ccanonkk + \Ckf~.
\eeq
The pointmap initially has larger errors and less confidence due to only using small baseline frames, but filtering merges information from many viewpoints.  We experimented with different ways of updating the canonical pointmap, and found that weighted average was best for maintaining coherence while filtering out noise.  Compared to the canonical pointmap in \mastersfm \cite{duisterhof_2024_mast3rsfm}, we compute this incrementally and require transformation of the points since an additional network prediction of $\Xkk$ would slow down tracking.  Filtering has a rich history in SLAM, and yields the benefit of leveraging information from all frames without having to explicitly optimise for all camera poses and store all predicted pointmaps from the decoder in the backend. 

\begin{table*}[t]
\centering
\caption{Absolute trajectory error (ATE (m)) on TUM RGB-D \cite{sturm_benchmark_2012}.}\label{tab:tum_ate}
\scriptsize
\begin{tabular}{l|lcccccccccc} 
& &\textbf{360} &\textbf{desk} &\textbf{desk2} &\textbf{floor} &\textbf{plant} &\textbf{room } &\textbf{rpy} &\textbf{teddy} &\textbf{xyz} &\textbf{avg} \\
\hline
\multirow{8}{*}{Calibrated} &\textbf{ORB-SLAM3 \cite{campos_orbslam3_2021}} &X &\underline{0.017} &0.210 &X &0.034 &X &X &X &\textbf{0.009} &- \\
&\textbf{DeepV2D \cite{teed_deepv2d_2020}} &0.243 &0.166 &0.379 &1.653 &0.203 &0.246 &0.105 &0.316 &0.064 &0.375 \\
&\textbf{DeepFactors \cite{czarnowski_deepfactors_2020}} &0.159 &0.170 &0.253 &0.169 &0.305 &0.364 &0.043 &0.601 &0.035 &0.233 \\
&\textbf{DPV-SLAM \cite{lipson2024deep}} &0.112 &0.018 &0.029 &0.057 &0.021 &0.330 &0.030 &0.084 &\underline{0.010} &0.076 \\
&\textbf{DPV-SLAM++ \cite{lipson2024deep}} &0.132 &0.018 &0.029 &0.050 &0.022 &0.096 &0.032 &0.098 &\underline{0.010} &0.054 \\
&\textbf{GO-SLAM \cite{zhang2023goslam}} &0.089 &\textbf{0.016} &\underline{0.028} &\underline{0.025} &0.026 &\underline{0.052} &\textbf{0.019} &0.048 &\underline{0.010} &\underline{0.035} \\
&\textbf{DROID-SLAM \cite{teed_droid_2021}} &0.111 &0.018 &0.042 &\textbf{0.021} &\textbf{0.016} &\textbf{0.049} &\underline{0.026} &0.048 &0.012 &0.038 \\
&\textbf{Ours} &\textbf{0.049} &\textbf{0.016} &\textbf{0.024} &\underline{0.025} &\underline{0.020} &0.061 &0.027 &\textbf{0.041} &\textbf{0.009} &\textbf{0.030} \\
\hline
\multirow{2}{*}{Uncalibrated} &\textbf{DROID-SLAM* \cite{teed_droid_2021, veicht2024geocalib}} &0.202 &0.032 &0.091 &0.064 &0.045 &0.918 &0.056 &\underline{0.045} &0.012 &0.158 \\
&\textbf{Ours*} &\underline{0.070} &0.035 &0.055 &0.056 &0.035 &0.118 &0.041 &0.114 &0.020 &0.060 \\
\bottomrule
\end{tabular}
\end{table*}

\subsection{Graph Construction and Loop Closure}

When tracking, a new keyframe $\KF_i$ is added if the number of valid matches or the number of unique keyframe pixels in $\match{f}{k}$ falls below a threshold $\KFthreshold$.  
After adding  $\KF_i$, a bidirectional edge to the previous keyframe  $\KF_{i-1}$is added to the edge-list $\mathcal{E}$.
This constrains the estimated poses sequentially in time; however, drift can still occur.  To close both small and large loops, we adapt the Aggregated Selective Match Kernel (ASMK) \cite{tolias_asmk_2013, tolias_learning_2020} framework used by \mastersfm~\cite{duisterhof_2024_mast3rsfm} for image retrieval from encoded features.  
While this was previously used in a batch setting where all images are available from the start, we modify it to work incrementally.  
We query the database with the encoded features of $\KF_i$ to obtain the top-$K$ images. Since the codebook only has tens of thousands of centroids, we found that a dense L2 distance calculation was sufficiently fast to quantise the features.  
If the retrieval scores are above a threshold $\Retrievalthreshold$, we give these pairs to the \master decoder and add bidirectional edges if the number of matches from \cref{sec:matching} is above a threshold $\LCthreshold$.  Lastly, we update the retrieval database by adding the new keyframe's encoded features to the inverted file index.

\subsection{Backend Optimisation}
Given current estimates of keyframe poses $\Twci$ and canonical pointmaps $\Xcanonii$ for $\KF_i$, the goal of the backend optimisation is to achieve global consistency across all poses and geometry.  While previous formulations used first-order optimisation and require rescaling after every iteration \cite{wang_dust3r_cvpr24, duisterhof_2024_mast3rsfm}, we introduce an efficient second-order optimisation scheme that handles the gauge freedom of the problem by fixing the first 7-DoF \Sim{3} pose.  We jointly minimise the ray error for all edges $\mathcal{E}$ in the graph:
\beq
E_g {=}\sum_{{i,j} \in \mathcal{E}}\sum_{{m, n} \in \match{i}{j}}\left\|\frac{\ray{\Xcanoniim} - \ray{\Tij \Xcanonjjn}}{w(\matchconf{m}{n}, \sigma^2_r)}\right\|_{\rho}, \label{eq:backend_ray}
\eeq
where $\Tij=\Twci^{-1}\Twcj$.
Given N keyframes, \cref{eq:backend_ray} forms and accumulates $14 \times 14$ blocks into the $7N\times 7N$ Hessian.  We solve this problem again using Gauss-Newton as in \cref{eq:gauss_newton} but with sparse Cholesky decomposition as the system is not dense.  Construction of the Hessian is made efficient through the use of analytical Jacobians and parallel reductions all implemented in CUDA.
Again, a small error term on consistency in distances is added to avoid degeneracy in the pure-rotation case. At most 10 iterations of Gauss-Newton are performed for every new keyframe and optimisation terminates upon convergence.  Second-order information greatly speeds up the global optimisation over the alternatives, and our efficient implementation ensures that it is not the bottleneck in the overall system.  

\subsection{Relocalisation}
If the system loses tracking due to an insufficient number of matches, relocalisation is triggered.  For a new frame, the retrieval database is queried with a stricter threshold on the score.  Once the retrieved images have a sufficient number of matches with the current frame, it is then added as a new keyframe into the graph and tracking resumes.

\subsection{Known Calibration}
Our system works without known camera calibration, but if we do have calibration we can make use of it to improve accuracy via two straightforward changes.  First, before canonical pointmaps are used for optimisation in both tracking and mapping, we query only the depth dimension and constrain the pointmap to be backprojected along the rays defined by the known camera model.  Second, we change the residuals in optimisation to be in pixel space rather than ray space.  In the backend, a pixel $\pix^i_{i,m}$ in $\Ii$ is compared against the projection of the 3D point it is matched with:
\beq
E_\projection=\sum_{{i,j} \in \mathcal{E}}\sum_{{m, n} \in \match{i}{j}}\left\|\frac{\pix^i_{i,m} - \proj{\Tij \Xcanonjjn}}{w(\matchconf{m}{n}, \sigma^2_\projection)}\right\|_{\rho}, \label{eq:backend_pixel}
\eeq
where $\projection$ is the projection function to pixel space using the given camera model.  Furthermore, the additional distance residuals are converted to depth for consistency.

\begin{figure}[t]
	\centering
	\includegraphics[width=0.98\columnwidth]{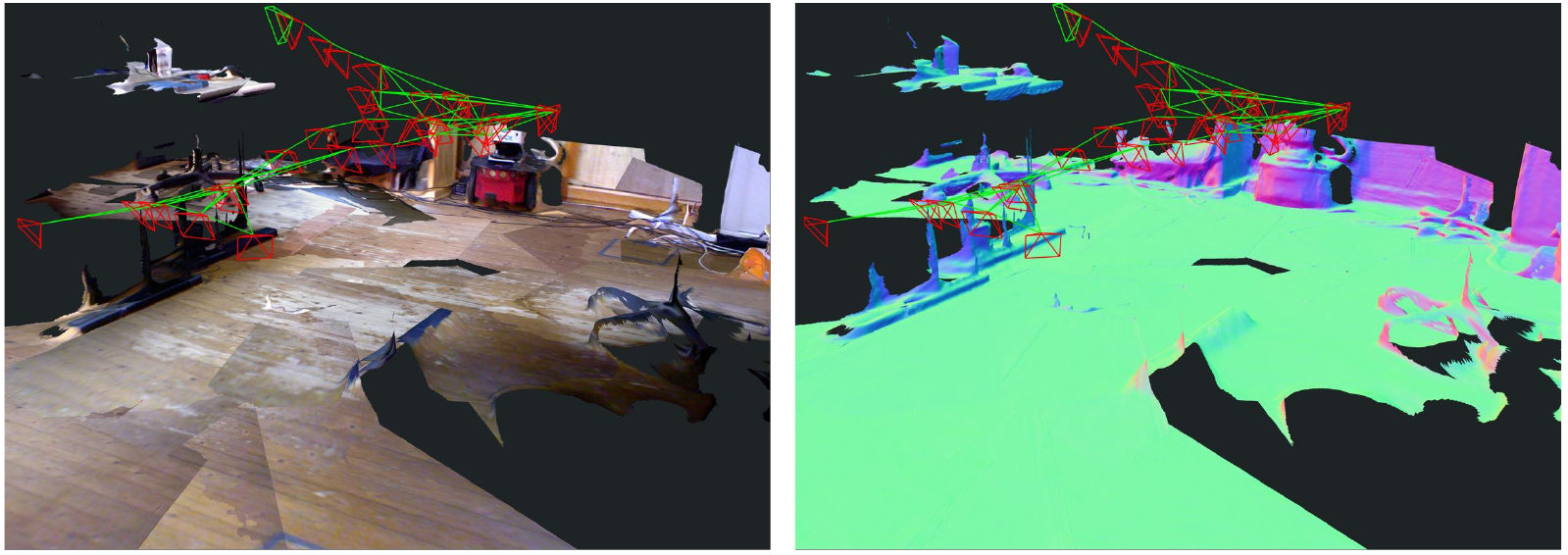}
	\caption{Reconstruction and trajectory TUM fr1/floor sequence.}
	\label{fig:tum_floor} 
\end{figure}

\section{Results}
\label{sec:results}
We evaluate our system on a wide range of real-world datasets. For localisation, we evaluate monocular SLAM on TUM RGB-D \cite{sturm_benchmark_2012}, 7-Scenes \cite{shotton_7scenes_2013}, ETH3D-SLAM \cite{schops_badslam_2019}, and EuRoC \cite{burri_euroc_2016}, all under monocular RGB setting. For geometry evaluation, we use the EuRoC Vicon room sequences as it provides 3D structure scan ground truth, as well as 7-Scenes since it has depth camera measurements.

We run our system on a desktop with Intel Core i9 12900K 3.50GHz and a single  NVIDIA GeForce RTX 4090.  As our system runs at roughly 15 FPS, we subsample every 2 frames of the datasets to simulate real-time performance.  Note that we use the full resolution outputs from \master, which resizes the largest dimension to size $512$.

\subsection{Camera Pose Estimation}\label{sec:camera-pose-estimation}
For all datasets, we report the RMSE of the absolute trajectory error (ATE) in metres.  Since all systems are monocular, we perform scaled trajectory alignment.  We denote our system without known calibration as Ours*.

\textbf{TUM RGB-D:} On the TUM dataset, we demonstrate state-of-the-art trajectory error when using calibration as shown in \cref{tab:tum_ate}.  Many of the previously best performing algorithms, such as DROID-SLAM, DPV-SLAM, and GO-SLAM, build on the foundational matching and end-to-end system proposed by DROID-SLAM.  In contrast, we propose a unique system that takes an off-the-shelf two-view geometric prior and show that it can outperform other systems while operating in real-time.  Furthermore, our uncalibrated system significantly outperforms a baseline, which we denote DROID-SLAM*, that calibrates the intrinsics using GeoCalib \cite{veicht2024geocalib} on the first image of a sequence, which is then used by DROID-SLAM.  We achieve this without assuming a fixed camera model across the entire sequence, and demonstrate the value of 3D priors for dense uncalibrated SLAM over priors that solve subproblems.  Our uncalibrated SLAM results are also comparable to results from recent learned techniques such as DPV-SLAM with known calibration.

\setlength{\tabcolsep}{2pt}
\begin{table}[t]
\centering
\caption{Absolute trajectory error (ATE (m)) on 7-Scenes \cite{shotton_7scenes_2013}.}\label{tab:7_scenes}
\scriptsize
\begin{tabular}{l|cccccccc} 
&\textbf{chess} &\textbf{fire} &\textbf{heads} &\textbf{office} &\textbf{pumpkin} &\textbf{kitchen} &\textbf{stairs} &\textbf{avg} \\
\hline
\textbf{NICER-SLAM} &\textbf{0.033} &0.069 &0.042 &0.108 &0.200 &\textbf{0.039} &0.108 &0.086 \\
\textbf{DROID-SLAM} &\underline{0.036} &\underline{0.027} &\underline{0.025} &\textbf{0.066} &0.127 &\underline{0.040} &\underline{0.026} &\underline{0.049} \\
\textbf{Ours} &0.053 &\textbf{0.025} &\textbf{0.015} &\underline{0.097} &\textbf{0.088} &0.041 &\textbf{0.011} & \textbf{0.047} \\
\textbf{Ours*} &0.063 &0.046 &0.029 &0.103 &\underline{0.114} &0.074 &0.032 &0.066 \\
\bottomrule
\end{tabular}
\end{table}

\textbf{7-Scenes:} We use the same sequences for evaluation following NICER-SLAM as shown in \cref{tab:7_scenes}.  Our calibrated system outperforms both NICER-SLAM \cite{zhu_nicer_2024} and DROID-SLAM.  Furthermore, our real-time uncalibrated system using a single 3D reconstruction prior outperforms NICER-SLAM, which uses multiple priors in depth, normal, and optical flow networks and runs offline.

\begin{figure}[t]
	\centering
	\includegraphics[width=0.98\columnwidth]{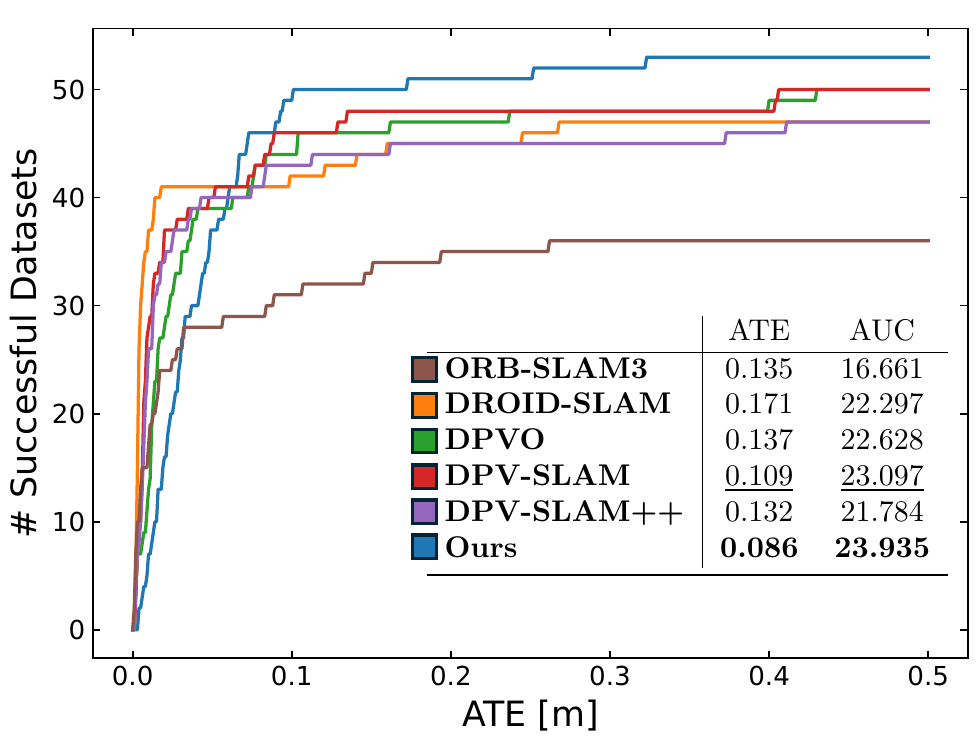}
	\caption{Number of successful trajectories below ATE threshold on ETH3D-SLAM (train) benchmark.  The corresponding table shows the mean ATE across completed sequences, as well as the AUC up to the threshold.}
	\label{fig:eth3d-train} 
\end{figure}

\textbf{ETH3D-SLAM:} Due to its difficulty, ETH3D-SLAM has only been evaluated for RGB-D methods.  Since the ATE thresholds for the official private evaluation are too strict for monocular methods, we evaluate several state-of-the-art monocular systems on the train sequences and generate the ATE curves.  The dataset contains sequences with fast camera motion, hence, for all methods, we do not subsample the frames. While other methods can have more precise trajectories, our method has a longer tail in terms of robustness, resulting in both the best ATE and area-under-curve (AUC).

\textbf{EuRoC:} We report the average ATE across all 11 EuRoC sequences in \cref{tab:geometry}. For the uncalibrated case, we found that the distortion was too significant as \master was not yet trained on such camera models, so we undistorted the images but did not give calibration to the rest of the pipeline. In general, our system is outperformed by DROID-SLAM, but it explicitly augments its training with 10\% greyscale images.  However, 0.041m ATE is still very accurate, and from the comparisons in \cite{lipson2024deep}, all outperforming methods build on top of the foundation from DROID-SLAM, while we present a novel method using a 3D reconstruction prior.

\subsection{Dense Geometry Evaluation}\label{sec:dense-geometry-eval}

We evaluate our geometry against DROID-SLAM and \spanner \cite{wang2024spann3r} on the EuRoC Vicon room sequences and 7-Scenes seq-01. 
For EuRoC, the alignment between the reference and the estimated point cloud is obtained by aligning the estimated trajectory against the Vicon trajectory.
Note, that this setup favours DROID-SLAM which obtains lower trajectory error.
For 7-Scenes, we backproject the depth images using poses provided by the dataset to create the reference point cloud. It is then aligned to the estimated point cloud using ICP as the extrinsic calibration between RGB and depth sensor is not provided. 

\begin{table}[t]\centering
\caption{Reconstruction Evaluation on 7-Scenes and EuRoC with all metrics in metres.}\label{tab:geometry}
\scriptsize
\begin{tabular}{l|cccc}
\textbf{7-scenes} &ATE &Accuracy &Completion &Chamfer \\
\hline
\textbf{DROID-SLAM} &\underline{0.049} & 0.115 & \textbf{0.040} & 0.077 \\
\textbf{\spanner@20} &N/A &\underline{0.069} &0.047 &\underline{0.058} \\
\textbf{\spanner@2} &N/A &0.124 &\underline{0.043} &0.084 \\
\textbf{Ours} &\textbf{0.047} & 0.074 & 0.057 & 0.066 \\
\textbf{Ours*} &0.066 & \textbf{0.068} &0.045 &\textbf{0.056} \\
\hline
\textbf{EuRoC} &ATE &Accuracy &Completion &Chamfer \\
\hline
\textbf{DROID-SLAM} &\textbf{0.022} &0.173 &\textbf{0.061} &0.117 \\
\textbf{Ours} &\underline{0.041} &\textbf{0.099} &\underline{0.071} &\textbf{0.085} \\
\textbf{Ours*} &0.164 &\underline{0.108} &0.072 &\underline{0.090} \\
\bottomrule
\end{tabular}
\end{table}

\begin{figure}[t]
\centering
\includegraphics[width=0.98\columnwidth]{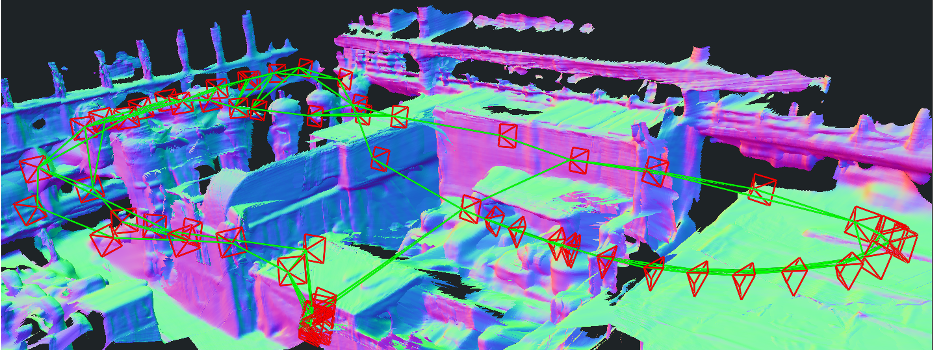}
\caption{Reconstruction on EuRoC Machine Hall 04.}
\label{fig:euroc_dense} 
\end{figure}

We report the RMSE for accuracy, which is defined as the distance between each estimated point and its nearest reference point, and completion, the distance between each reference point and its nearest estimated point. 
Both metrics are calculated with a maximum distance threshold of 0.5m and averaged across all sequences. 
We also report Chamfer Distance, the average of the two metrics. 

\cref{tab:geometry} summarises the geometry evaluation on 7-Scenes and EuRoC. For 7-Scenes, both our method with and without calibration and \spanner achieve more accurate reconstruction compared to DROID-SLAM, highlighting the advantage of the 3D prior. We run \spanner under two different settings. In one, a keyframe is taken every 20 images and in the other every 2 images. The discrepancy in the two settings shows the challenges test-time optimisation-free approaches face to generalise. Ours without calibration performs the best in both Accuracy and Chamfer distance. This can be attributed to the fact that the intrinsic calibration 7-Scenes provides is the default factory calibration.

For EuRoC, \spanner struggles as the sequences are not object-centric and thus is excluded. As summarised in \cref{tab:geometry}, although DROID-SLAM outperforms our method in terms of ATE, our method with/without calibration obtains better geometry.  DROID-SLAM obtains higher completion as it estimates a large number of noisy points which surround the reference point cloud, but our method has significantly better accuracy.  It is interesting to note that our uncalibrated system has a noticeably larger ATE, but still outperforms DROID-SLAM in Chamfer distance.  

\subsection{Qualitative Results}
\begin{figure}[t]
	\centering
	\includegraphics[width=0.98\columnwidth]{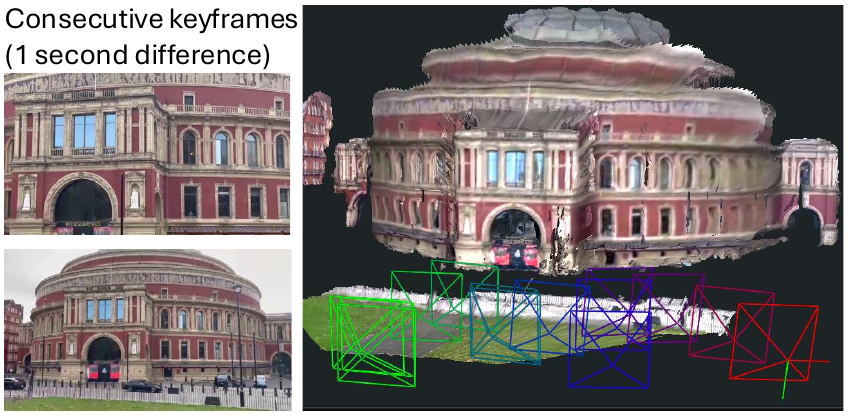}
	\caption{Dense uncalibrated SLAM with extreme zoom changes shown by two consecutive keyframes for an outdoor scene.}
	\label{fig:zoom_example} 
\end{figure}

\cref{fig:teaser} shows a reconstruction of the challenging Burghers sequence which has few matchable features on the specular figures.  We show examples of pose estimation and dense reconstructions for TUM in \cref{fig:tum_floor} and for EuRoC in \cref{fig:euroc_dense}.    Furthermore, we show an example with extreme zoom changes between consecutive keyframes in \cref{fig:zoom_example}.

\subsection{Component Analysis}

We compare matching techniques in \cref{tab:matching}.  Our parallelised projective matching with feature refinement achieves the best accuracy with significantly faster runtime.  Performing \master matching over all pixels takes 2 seconds, while our matching takes 2ms and makes the entire system FPS nearly 40x faster.  Please refer to the supplementary for a full runtime analysis of the system.  In \cref{tab:ablation_pointmap_fusion}, we test different methods for updating the canonical pointmap and report the average ATE across TUM, 7-Scenes, and EuRoC.  Selecting the most recent and first pointmaps incur drift and lack sufficient baseline, respectively.  Given calibration, weighted fusion performs on par with selecting the pointmap with the highest median confidence, but it achieves the lowest ATE without calibration and improves the ATE on EuRoC by 1.3cm, indicating that fusing over camera models is important.  In \cref{tab:ablation_point_ray}, the ray error formulation for uncalibrated tracking and backend optimisation improves performance over using the 3D point error which contains inaccurate depth predictions. \cref{tab:loop_closure} shows that loop closure improves both pose and geometry accuracy, with more significant gains on longer sequences. This demonstrates that the outputs of MASt3R still contain bias and cause drift, which our components are designed to mitigate.

\begin{table}[!t]
    \centering
    \setlength{\tabcolsep}{1pt}
    \begin{minipage}[t]{0.6\linewidth}
        \caption{Matching comparison.}\label{tab:matching}
        \scriptsize
        \begin{tabular}{l|cccc} 
        &ATE [m] &ATE [m] & Matching    &System \\
        &w/ calib &w/o calib & Time [ms] &FPS \\
        \hline
        \kdtree &0.061 &0.115 &40 &8.8 \\
        \master &\underline{0.042} &0.098 &2000 &0.4 \\
        Ours &0.062 &\textbf{0.092} &\textbf{0.5} &\textbf{15.1} \\
        Ours + feat &\textbf{0.039} &\underline{0.097} &\underline{2} &\underline{14.9} \\
        \bottomrule
        \end{tabular}
    \end{minipage}%
    \begin{minipage}[t]{0.4\linewidth}
        \centering
    \caption{Fusion methods.}\label{tab:ablation_pointmap_fusion}
    \scriptsize
    \begin{tabular}{l|cc} 
&ATE [m] &ATE [m] \\
    &w/o calib &w/ calib \\
    \hline
    Recent &0.207 &0.160 \\
    First &0.114 &\underline{0.059} \\
    Median &\underline{0.102} &\textbf{0.039} \\
    Weighted &\textbf{0.097} &\textbf{0.039} \\
    \bottomrule
    \end{tabular}
    \end{minipage}
\end{table}

\begin{table}[!t]
    \centering
        \caption{ATE (m) for error formulation in format (point / ray).}\label{tab:ablation_point_ray}
        \scriptsize
        \begin{tabular}{ccc|c} 
        TUM &7-Scenes &EuRoC &avg \\
        \hline
        0.092 / \textbf{0.060} &0.084 / \textbf{0.066} &0.290 / \textbf{0.164} &0.155 / \textbf{0.097}\\
        \bottomrule
        \end{tabular}
\end{table}

\begin{table}[!t]\centering
\caption{Loop closure ablation in format (without LC / with LC).} \label{tab:loop_closure}
\scriptsize
\begin{tabular}{l|ccc|c}
& \multicolumn{3}{c|}{ATE (m)} &Chamfer (m) \\
& TUM &7-scenes & EuRoC Vicon & EuRoC Vicon \\ \hline
Calib & 0.064 / \textbf{0.030} &0.066 / \textbf{0.047} &0.233 / \textbf{0.029} &0.151 / \textbf{0.085} \\
No Calib & 0.090 / \textbf{0.060} &0.075 / \textbf{0.066} &0.349 / \textbf{0.122} &0.179 / \textbf{0.090} \\
\bottomrule
\end{tabular}
\end{table}

\section{Limitations and Future Work}
\label{sec:limitations}

While we can estimate accurate geometry by filtering pointmaps in the frontend, we do not currently refine all geometry in the full global optimisation.  While DROID-SLAM optimises per-pixel depth via bundle adjustment, this framework permits incoherent geometry.  A method that can make pointmaps globally consistent in 3D while retaining the coherence of the original \master predictions all in \textit{real-time} would be an interesting direction for future work.  

Since \master is only trained on images with pinhole images, its geometry predictions degrade with increasing distortion.  However, in the future, models will be trained on a variety of camera models and will be compatible with our framework that never assumes a parametric camera model.  Furthermore, using the decoder at full resolution is currently a bottleneck, especially for low-latency tracking and checking loop closure candidates.  Improving network throughout will benefit the total system efficiency. 

\section{Conclusion}
\label{sec:conclusion}
We present a real-time dense SLAM system based on \master that handles in-the-wild videos and achieves state-of-the-art performance.  Much of the recent progress in SLAM has followed the contributions of DROID-SLAM, which trains an end-to-end framework that solves for poses and geometry from a flow update. 
We take a different approach by building a system around an off-the-shelf geometric prior that achieves comparable pose estimation for the first time, while also providing consistent dense geometry.

\section{Acknowledgement}
This research is supported by the Engineering and Physical
Sciences Research Council [grant number EP/W524323/1].
{
    \small
    \bibliographystyle{ieeenat_fullname}
    \bibliography{main}
}

\clearpage
\maketitlesupplementary

\section{Analytical Jacobians}

In this section, we derive analytical Jacobians used in second-order optimisation for both the tracking and backend.  For more information on Lie algebra and relevant Jacobians, please see the following \cite{sola_micro_2018, teed2021tangent}.

To take the derivatives on Lie groups with respect to the minimal parameterisation, we use the left-Jacobian definition:
\bea
\frac{\cD f(\matT)}{\cD \matT} &\triangleq \lim_{\vectau \rightarrow 0} \frac{f(\vectau \oplus \matT) \ominus f(\matT)}{\vectau}~,\\
&= \lim_{\vectau \rightarrow 0} \frac{\text{Log}\left(f\left(\text{Exp}(\vectau)\circ \matT\right) \circ f(\matT)^{-1}\right)}{\vectau}.
\eea

\subsection{Points}

For the point alignment used in both tracking and mapping, we have a residual defined between a measured point in one frame and a transformed point matched from a different frame. Using the general notation from the backend for point alignment, and switching the order of the residual which does not affect the cost function, the residual is: 
\beq
r_p = \Tij \Xcanonjjn - \Xcanoniim.
\eeq 
Defining $\vecx = \Tij \Xcanonjjn$ for brevity in deriving Jacobians for a single point, we take the partial derivatives with respect to the Lie algebra perturbation of the relative pose $\Tij$:
\beq
    \frac{\cD r_p}{\cD \Tij} = \begin{bmatrix} \mbf{I}_{3\times 3} & -[\vecx]_{\times} & \vecx \end{bmatrix}
    \label{eq:point_jac}
\eeq
where $[\vecx]_{\times}$ is the $3\times3$ skew-symmetric matrix.

\subsection{Rays and Distance}

Compared to the point residual, the ray residual minimises the error in normalised space, which is equivalent to minimising the angle between rays in the camera's frame:
\beq
r_\psi = \ray{\Tij \Xcanonjjn} - \ray{\Xcanoniim}.
\eeq
The Jacobian now is the chain rule of the Jacobian for normalising a point to a unit vector and Jacobian of the the pose acting on the point:
\beq
    \frac{\cD _\psi}{\cD \Tij} = \pdiff{r_\psi}{\vecx} \frac{\cD \vecx}{\cD \Tij}.
\eeq
Defining the distance from the origin of camera $i$ to point $\vecx$ as $d_{\vecx}$, the Jacobian of the first term becomes: 
\beq
    \pdiff{r_\psi}{\vecx} = \frac{1}{d_{\vecx}} \left( \mbf{I}_{3\times 3} - \frac{\vecx \vecx^T}{d_{\vecx}^2} \right).
    \label{eq:norm_jac}
\eeq
Using the chain rule with \cref{eq:point_jac}, the first term becomes \cref{eq:norm_jac} itself.  Since the cross product of a point with itself is a zero vector, the second term becomes a scaled version of the skew-symmetric matrix.  Lastly, as \cref{eq:norm_jac} has the form of an operator that takes the difference between a point and its orthogonal projection onto a subspace, and projecting a point onto its own subspace preserves the point, this cancels to a zero vector.  In matrix form, this is:
\beq
    \frac{\cD r_\psi}{\cD \Tij} = \begin{bmatrix} \pdiff{r_\psi}{\vecx} & -\frac{1}{d_{\vecx}} [\vecx]_{\times} & \mbf{0}_{3 \times 1} \end{bmatrix}.
\eeq
As mentioned in the main paper, we also include an error based on the distance between the transformed point and its match so that cases with pure rotation do not result in a degenerate optimisation problem.  This error is:
\beq
    r_d = d\left(\Tij \Xcanonjjn \right) - d\left( \Xcanoniim \right)
\eeq
and its corresponding Jacobians are:
\bea
    \pdiff{r_d}{\vecx} &= \frac{\vecx^T}{d_{\vecx}}, \\
    \frac{\cD r_d}{\cD \Tij} &= \begin{bmatrix} \frac{\vecx^T}{d_{\vecx}} & \mbf{0}_{1 \times 3} & d_{\vecx} \end{bmatrix}.
    \label{eq:dist_jac}
\eea

\begin{figure*}[t]
\centering
\includegraphics[width=\linewidth]{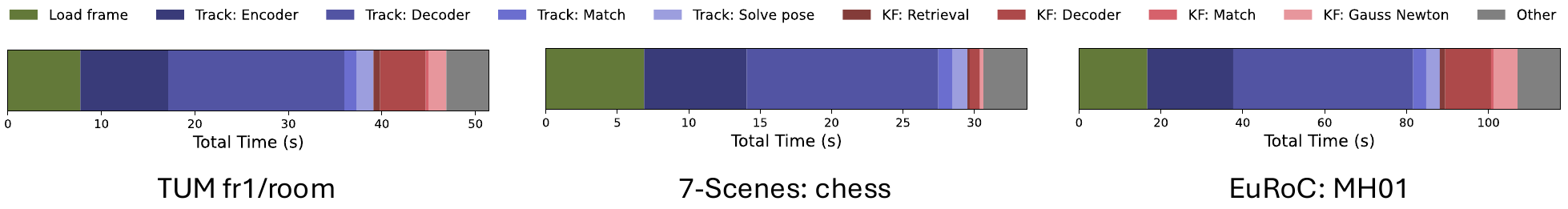}
\caption{Total runtime in seconds for representative datasets showing cumulative time spent in significant components.  The network encoder and decoder are the majority of the runtime at an average of 64\% of the total runtime.  Datasets with more loop closures like fr1/room and MH01 show more time spent in the backend.}
\label{fig:timing_results}
\end{figure*}

\begin{table*}[t]\centering
\caption{Average runtimes in milliseconds of different components for our single-threaded system.}\label{tab:timing}
\footnotesize
\begin{tabular}{l V{3} c V{2} cccc|c V{2} cccc|c V{3} cc}
& Data &\multicolumn{5}{c V{2}}{Per-Frame Tracking} &\multicolumn{5}{c V{3}}{Per Keyframe} & \multicolumn{2}{c}{Summary} \\
    &\multicolumn{1}{c V{2}}{Load frame} &Encoder &Decoder &Match &Solve pose &\multicolumn{1}{c V{2}}{Total} &Retrieval &Decoder &Match &Gauss-Newton &\multicolumn{1}{c V{3}}{Total} & Total time (s) & FPS \\
\hline
TUM: fr1/room &11.4 &13.8 &27.7 &1.9 &2.7 &48.6 &14.4 &\ \ 97.7 &5.8 &37.8 &157.4 &51.5 &13.2 \\
7-Scenes: chess &13.8 &14.4 &26.9 &2.0 &2.1 &47.9 &14.4 &\ \ 70.6 &4.5 &22.7 &114.0 &33.7 &14.8 \\
EuRoC: MH01 &\ \ 9.1 &11.4 &23.9 &1.7 &1.8 &41.1 &15.1 &130.4 &8.4 &66.8 &223.3 &117.6 &15.7 \\
\hline
Average &11.4 &13.2 &26.2 &1.9 &2.2 &45.9 &14.6 &\ \ 99.5 &6.2 &42.4 &164.9 &67.6 &14.6 \\
\bottomrule
\end{tabular}
\end{table*}

\subsection{Projection and Depth}

In the case of known calibration, we instead use a pixel error instead of ray error.  While the rays could also be constrained to the known camera model, we chose to use pixel error as this better models the noise distribution in pixel-level correspondence and is standard in bundle adjustment.  The pixel error is defined as:
\beq
    r_\projection = \proj{\Tij \Xcanonjjn} - \pix^i_{i,m}.
\eeq
Using a pinhole camera model with calibration
\beq
    K = \begin{bmatrix} f_x & 0 & c_x \\ 0 & f_y & c_y \\ 0 & 0 & 1 \end{bmatrix},
\eeq
the projection Jacobian of point $\vecx = [x,y,z]^T$ is
\beq
    \pdiff{r_\projection}{\vecx} = \frac{1}{z} \begin{bmatrix} f_x & 0 & -f_x \frac{x}{z} \\ 0 & f_y & -f_y \frac{y}{z} \end{bmatrix}.
\eeq
We can then obtain $\frac{\cD r_\projection}{\cD \Tij}$ via the chain rule with \cref{eq:point_jac}.  We also include a small error on the predicted and measured depth with similar motivation to \cref{eq:dist_jac} in cases of pure rotation.  In the future, any parametric camera model and its corresponding Jacobian could be used here.

\subsection{From Relative Pose to Global Pose}

While the above derivations show the Jacobians with respect to relative camera poses, we ultimately need updates with respect to camera poses in the world frame.
Using $\Tij=\Twci^{-1}\Twcj$ and the identities for the left Jacobian of the group inverse and composition
\bea
    \frac{\cD \Twci^{-1}}{\cD \Twci} &= -\text{Ad}_{\Twci^{-1}}, \\
    \frac{\cD \Tij}{\cD \Twci^{-1}} &= \mbf{I}_{7\times 7}, \\
    \frac{\cD \Tij}{\cD \Twcj} &= \text{Ad}_{\Twci^{-1}},
\eea
we can then solve for updates to each pose: 
\bea
    \frac{\cD r_\psi}{\cD \Twci} &= \frac{\cD r_\psi}{\cD \Tij} \frac{\cD \Tij}{\cD \Twci} = - \frac{\cD r_\psi}{\cD \Tij} \text{Ad}_{\Twci^{-1}}, \\
    \frac{\cD r_\psi}{\cD \Twcj} &= \frac{\cD r_\psi}{\cD \Tij} \frac{\cD \Tij}{\cD \Twcj} = \frac{\cD r_\psi}{\cD \Tij} \text{Ad}_{\Twci^{-1}}.
\eea

\section{Initialisation}
As mentioned in \cref{sec:tracking}, to minimise the number of network passes required for tracking, we re-use the last keyframe's pointmap estimate $\Xcanonkk$.
Such pointmap is always available, apart from at the initialisation.
To initialise the system, we simply feed the same image into \master to perform monocular prediction of the pointmap. 
While such monocular predictions are often inaccurate, the pointmap incorporates multiview information and is refined using the running weighted average filter.

\section{Runtime Breakdown}

We report the cumulative runtime for different components of our system across three representative datasets in \cref{fig:timing_results}.  We also show average runtimes of different components in \cref{tab:timing}.  Note that tracking, which operates at greater than 20 FPS, occurs for every frame while keyframing is dependent on the motion and thus occurs at a lower frequency.  In general, the network encoder and decoder are the most significant in terms of time spent for both the tracking and backend at around 64\% of the total runtime.  As a large number of loop closures are detected in TUM fr1/room and EuRoC MH01, the time spent in the backend increases compared to the more linear trajectory in 7-Scenes chess.  Our efficient matching, tracking, and backend optimisation ensure that we can achieve real-time performance, with the network currently being the limiting factor on lower-latency SLAM.  The combination of the modular prior and principled backend optimisation achieves global consistency in real-time. 

\section{Evaluation Setup}

\begin{figure*}[t]
    \centering
    \includegraphics[width=\textwidth]{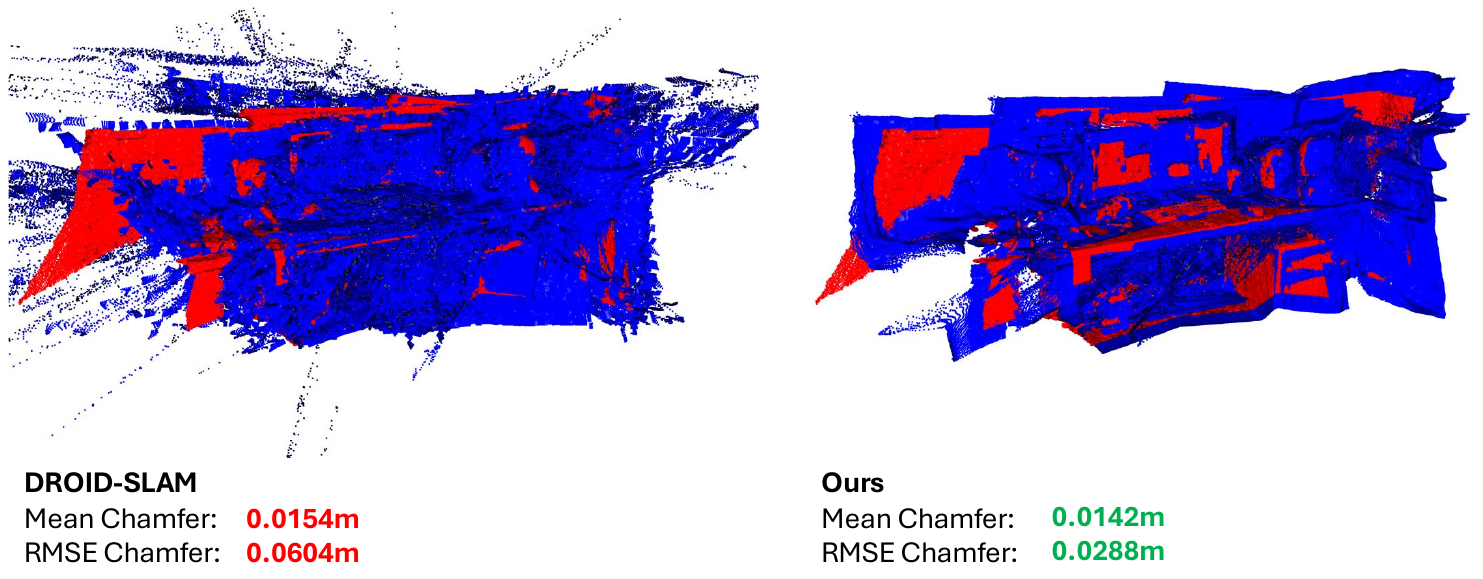}
    \caption{Reconstruction comparison on 7-Scenes heads, with red indicating the ground-truth point cloud and blue the estimated point cloud.  While mean Chamfer distance does not significantly penalise inconsistent points, RMSE Chamfer is a better reflection of the quality of the geometry.}
    \label{fig:7scenes_geometry} 
\end{figure*}

\begin{figure*}[t]
    \centering
    \includegraphics[width=\textwidth]{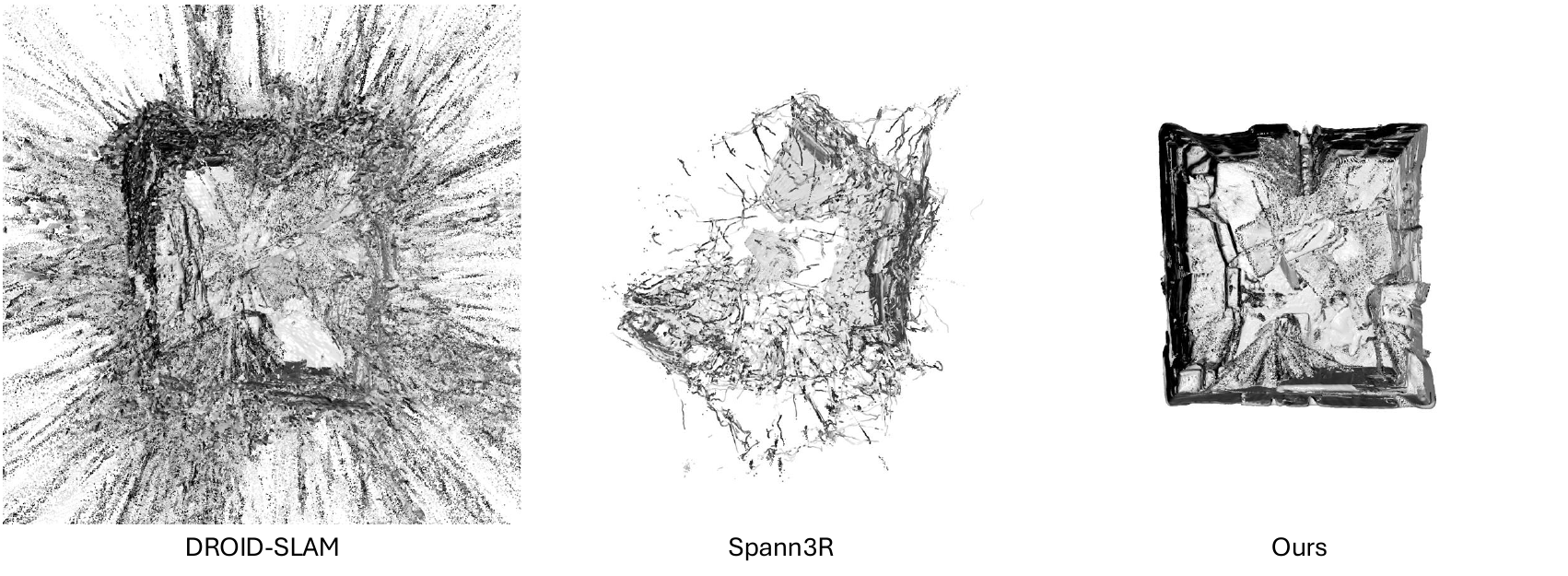}
    \caption{Reconstruction comparison on EuRoC V102.}
    \label{fig:euroc_geometry} 
\end{figure*}

\subsection{Trajectory Evaluation [\cref{sec:camera-pose-estimation}]]}
For all the datasets, we use the same parameters with keyframe threshold $\KFthreshold = 0.333$, loop-closure threshold $\LCthreshold= 0.1$, and $\Retrievalthreshold=0.005$.  
For relocalisation, we have a stricter check to allow for the current frame to be attached to the graph.  The match fraction must be greater than 0.3 for all datasets apart from in ETH3D where we set the threshold higher to 0.5.

For trajectory evaluation, we run DROID-SLAM using the open-source code with the configuration files given for each dataset.  For 7-Scenes, we use the TUM configuration file since it is the most similar.  For TUM and EuRoC, the remaining entries are from the tables in Deep Patch Visual SLAM \cite{lipson2024deep}, which also uses some results from DROID-SLAM \cite{teed_droid_2021}.  For 7-Scenes, we include the results reported from NICER-SLAM \cite{zhu_nicer_2024}.  For ETH3D, we ran all methods locally as the dataset was not previously attempted with monocular SLAM methods.

\subsection{Geometry Evaluation [\cref{sec:dense-geometry-eval}]}
For evaluation, points that are unobservable are removed from the reference point cloud.
Additionally, for the 7-Scenes dataset, we filter out depths which are marked as invalid.
For all methods, we do not filter any estimated point, as in an incremental problem setting like SLAM, reprojection-based filtering is not always possible and downstream applications benefit from per-pixel dense prediction.

For the metrics, we report the RMSE which penalises outlying measurements. 
\cref{fig:7scenes_geometry} is an illustrative example, where DROID-SLAM and \master-SLAM achieve a similar mean Chamfer distance.  Qualitatively, however, \master-SLAM clearly produces more coherent and accurate geometry, and this difference is reflected in the RMSE Chamfer distance.

\begin{table*}[t]\centering
\caption{Absolute trajectory error (ATE (m)) on EuRoC \cite{burri_euroc_2016}.}\label{tab:euroc_ate}
\footnotesize
\begin{tabular}{l|lcccccccccccc} 
& &MH01 &MH02 &MH03 &MH04 &MH05 &V101 &V102 &V103 &V201 &V202 &V203 &avg \\
\hline
\multirow{8}{*}{Calibrated} &\textbf{ORB-SLAM} &0.071 &0.067 &0.071 &0.082 &0.060 &\textbf{0.015} &0.020 &X &0.021 &0.018 &X &- \\
&\textbf{DeepV2D \cite{teed_deepv2d_2020}} &0.739 &1.144 &0.752 &1.492 &1.567 &0.981 &0.801 &1.570 &0.290 &2.202 &2.743 &1.298 \\
&\textbf{DeepFactors \cite{czarnowski_deepfactors_2020}} &1.587 &1.479 &3.139 &5.331 &4.002 &1.520 &0.679 &0.900 &0.876 &1.905 &1.021 &2.040 \\
&\textbf{DPV-SLAM \cite{lipson2024deep}} &\textbf{0.013} &0.016 &\underline{0.022} &\underline{0.043} &\textbf{0.041} &\underline{0.035} &\textbf{0.008} &\textbf{0.015} &0.020 &\underline{0.011} &0.040 &0.024 \\
&\textbf{DPV-SLAM++ \cite{lipson2024deep}} &\textbf{0.013} &0.016 &\textbf{0.021} &\textbf{0.041} &\textbf{0.041} &\underline{0.035} &\underline{0.010} &\textbf{0.015} &0.021 &\underline{0.011} &0.023 &\underline{0.023} \\
&\textbf{GO-SLAM \cite{zhang2023goslam}} &\underline{0.016} &\underline{0.014} &0.023 &0.045 &0.045 &0.037 &0.011 &0.023 &\textbf{0.016} &\textbf{0.010} &\underline{0.022} &0.024 \\
&\textbf{DROID-SLAM \cite{teed_droid_2021}} &\textbf{0.013} &\textbf{0.012} &\underline{0.022} &0.048 &\underline{0.044} &0.037 &0.013 &\underline{0.019} &\underline{0.017} &\textbf{0.010} &\textbf{0.013} &\textbf{0.022} \\
&\textbf{Ours} &0.023 &0.017 &0.057 &0.113 &0.067 &0.040 &0.019 &0.027 &0.020 &0.025 &0.043 &0.041 \\
\hline
Uncalibrated &\textbf{Ours*} &0.180 &0.124 &0.156 &0.282 &0.327 &0.101 &0.134 &0.096 &0.133 &0.100 &0.170 &0.164 \\
\bottomrule
\end{tabular}
\end{table*}

We report the qualitative result of EuRoC reconstruction in \cref{fig:euroc_geometry}. \spanner fails as the sequence is not object-centric, and DROID-SLAM produces many more outliers compared to \master-SLAM .  Compared to \spanner which maintains a memory buffer, our keyframing system ensures that viewed parts of the scene are not discarded.  Furthermore, our efficient global optimisation can create globally consistent maps in real-time.

\section{EuRoC Results}

We summarise the average ATE for EuRoC in the main paper, and show the results for each sequence in \cref{tab:euroc_ate}.  While our system does not outperform DROID-SLAM and methods that leverage its matching architecture, EuRoC has traditionally been challenging for monocular systems due to aggressive motion, large-scale trajectories, and varying exposure.  As noted previously, DROID-SLAM was trained with explicit greyscale augmentation which may account for the gap in performance.  Compared to previous systems with geometric priors, such as DeepV2D and DeepFactors, we demonstrate significant improvements in trajectory estimation.  Furthermore, the results from the main paper highlight the additional benefits of using such a prior, as the dense geometry is more accurate and consistent as shown in \cref{tab:geometry}, even for our uncalibrated system.

\section{Comparison to Other SLAM/SfM Methods}

\subsection{DROID and DPV SLAM}

Our system uses a two-view geometric prior in a modular system, while DROID and DPV SLAM learn a matching prior as part of an end-to-end system with differentiable bundle adjustment.  While these systems are very accurate for pose estimation, there are fundamental limitations for geometry and generality.

First, bundle adjustment cannot guarantee coherent geometry even with accurate poses, as it lacks smoothness regularisation and constraints under low parallax.  Given MASt3R, we find local fusion and scale optimisation to be sufficient for consistency and coherence, while the BA of DROID loses the latter. 
Second, beyond improved geometry, geometric priors enable new capabilities like continuously changing intrinsics.  DROID fixes the model to pinhole during training, and also cannot efficiently handle time-varying intrinsics as this slows down backend optimisation.

\subsection{MASt3R-SfM}

MASt3R-SfM uses sparse correspondences (subsampling 1/64 pixels) due to MASt3R's brute-force matching.  Our dense projective pointmap matching formulates search as local optimisation and achieves a 1000x speedup without compromising accuracy as shown in \cref{tab:matching}. Global optimisation in MASt3R-SfM uses a 1st-order optimiser, lacks minimal rotation updates, and introduces degenerate solutions that require scale renormalisation.
To avoid such problems, we formulate a nonlinear least-squares problem and develop a 2nd-order optimiser with minimal pose updates and gauge fixing.  
Our uncalibrated ray formulation achieves a similar accuracy as MASt3R-SfM's procedure of fitting pinhole models and minimising reprojection error, but we avoid selecting a specific camera model. This maintains generality of our SLAM system in order to handle all types of distortion, such as fisheye in the future.


\end{document}